\documentclass[letterpaper, 10 pt, conference]{ieeeconf}  % Comment this line out if you need a4paper

\IEEEoverridecommandlockouts                              % This command is only needed if 
\usepackage{graphicx} % for pdf, bitmapped graphics files
\usepackage{amsmath} % assumes amsmath package installed
\usepackage{amssymb}  % assumes amsmath package installed
\usepackage{algorithm}
\usepackage{algorithmic}
\usepackage{booktabs}       % professional-quality tables
\usepackage{multirow}  % for tables
\usepackage{makecell}  % for tables
\usepackage{url}
\title{\LARGE \bf
 CenterRadarNet: Joint 3D Object Detection and Tracking Framework using 4D FMCW Radar
}

\author{Jen-Hao Cheng$^{1}$, Sheng-Yao Kuan$^{2}$, Hugo Latapie$^{3}$, Gaowen Liu$^{3}$, Jenq-Neng Hwang$^{1}$ 
\thanks{$^{1}$Jen-Hao Cheng and Jenq-Neng Hwang are with University of Washington, USA. \tt\small \{andyhci, hwang\}@uw.edu
        }
\thanks{$^{2}$Sheng-Yao Kuan is with National Yang Ming Chiao Tung University, Taiwan. \tt\small shaunkuan.10@nycu.edu.tw
        }
\thanks{$^{3}$Hugo Latapie and Gaowen Liu are with Cisco, USA. \tt\small \{hlatapie, gaoliu\}@cisco.com
        }
}

\begin{document}

\maketitle
\thispagestyle{empty}
\pagestyle{empty}

%%%%%%%%%%%%%%%%%%%%%%%%%%%%%%%%%%%%%%%%%%%%%%%%%%%%%%%%%%%%%%%%%%%%%%%%%%%%%%%%
\begin{abstract}
Robust perception is a vital component for ensuring safe autonomous and assisted driving. Automotive radar (77 to 81 GHz), which offers weather-resilient sensing, provides a complementary capability to the vision- or LiDAR-based autonomous driving systems. Raw radio-frequency (RF) radar tensors contain rich spatiotemporal semantics besides 3D location information. The majority of previous methods take in 3D (Doppler-range-azimuth) RF radar tensors, allowing prediction of an object’s location, heading angle, and size in bird’s-eye-view (BEV). However, they lack the ability to at the same time infer objects' size, orientation, and identity in the 3D space. 
To overcome this limitation, we propose an efficient joint architecture called CenterRadarNet, designed to facilitate high-resolution representation learning from 4D (Doppler-range-azimuth-elevation) radar data for 3D object detection and re-identification (re-ID) tasks. 

As a single-stage 3D object detector, CenterRadarNet directly infers the BEV object distribution confidence maps, corresponding 3D bounding box attributes, and appearance embedding for each pixel. Moreover, we build an online tracker utilizing the learned appearance embedding for re-ID. CenterRadarNet achieves the state-of-the-art result on the K-Radar 3D object detection benchmark. In addition, we present the first 3D object-tracking result using radar on the K-Radar dataset V2. In diverse driving scenarios, CenterRadarNet shows consistent, robust performance, emphasizing its wide applicability.

\end{abstract}

%%%%%%%%%%%%%%%%%%%%%%%%%%%%%%%%%%%%%%%%%%%%%%%%%%%%%%%%%%%%%%%%%%%%%%%%%%%%%%%%
\section{INTRODUCTION}
%% intro 
Safe autonomous driving and advanced driver assistance systems (ADAS) rely on resilient object perception under various driving conditions. To achieve robust 3D object detection and multiple object tracking (MOT), various sensors (e.g., cameras, LiDAR, radar) and data representations have been deployed and combined. 

\begin{figure}[t]
    \centering
    \includegraphics[width=\linewidth]{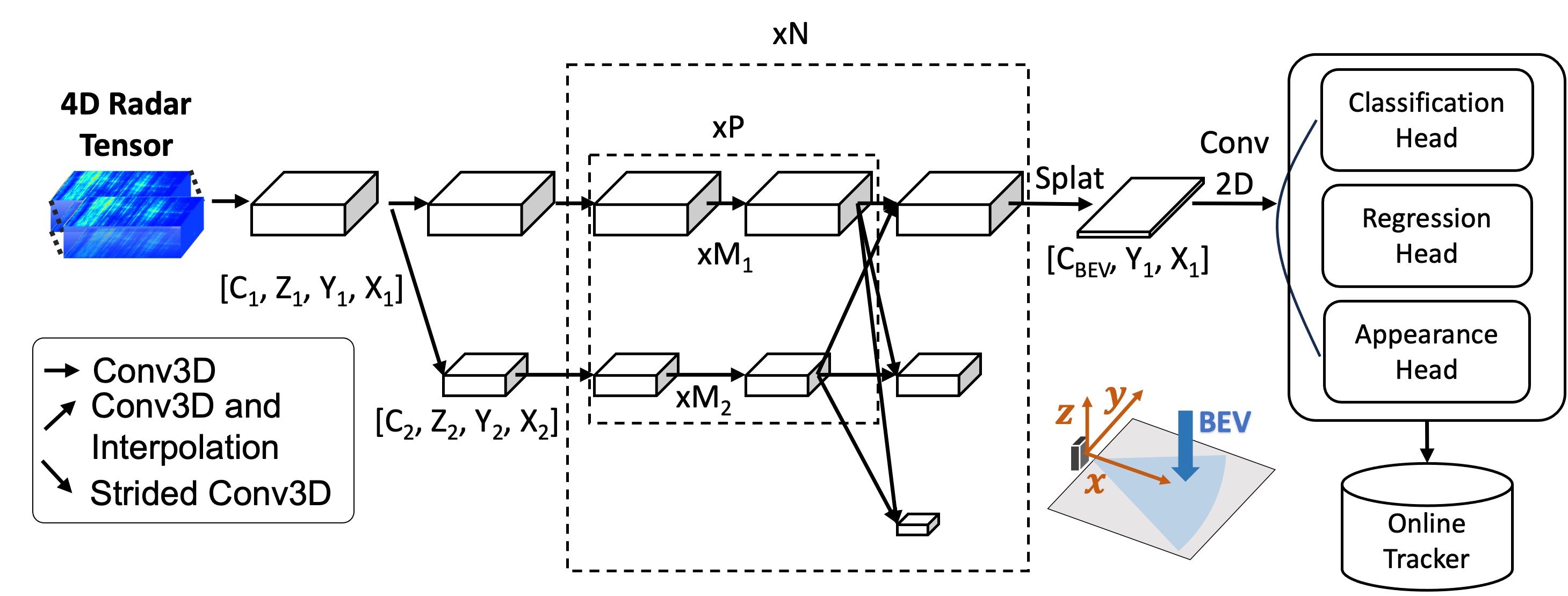}
    \caption{CenterRadarNet is a comprehensive framework designed for 3D object detection and tracking, specifically utilizing 4D FMCW radar tensors as its input. The architecture is comprised of three main components: a Conv3D backbone for feature extraction, specialized object detection and appearance heads for accurate object recognition, and an online tracker.}
    \label{fig:cenrad}
\end{figure}

% data representation (radar point cloud - radar tensor) comparison
In recent studies, radar has been deployed to compensate ADAS for camera and LiDAR under rigorous weather conditions in multiple forms, including raw radio-frequency (RF) radar tensors and their signal processing results, which are called radar point clouds.
While radar point cloud serves a sparse visual representation of objects, raw radio-frequency (RF) radar tensors preserve spatiotemporal information. 
Some existing works \cite{dong2020probabilistic, RADTensors,ouaknine2021carrada, wang2021rodnet2} take in 3D (Doppler-range-azimuth) RF radar tensors as input and predict object information in the bird's-eye view (BEV).
However, these methods rely on 3D radar tensors lacking elevation-wise information and do not learn appearance features for the re-identification (re-ID) task.
Therefore, we propose a novel framework, CenterRadarNet, using 4D (Doppler-range-azimuth-elevation) radar tensors to jointly predict objects' 3D bounding boxes and appearance features for the re-ID task and online tracking as illustrated in Fig. \ref{fig:cenrad}. 
We allow target detection and appearance embedding to be learned jointly. 
With CenterRadarNet's capability, we believe that radar can better assist other sensors in 3D perception.

The architecture of CenterRadarNet consists of a 3D backbone to learn spatiotemporal representation and a subsequent set of classification and regression heads for object detection and appearance embedding for re-ID tasks. 
Inspired by the success of HRNet~\cite{HRNet} on image perception tasks, we carefully design a 3D backbone called HR3D, which simultaneously preserves the high-resolution representation and aggregates information between multiple scales during the forward pass of the backbone network.
Consequently, the learned representation is favorable for detecting objects.

We extend CenterPoint~\cite{CenterPoint} first stage's heads to jointly detect objects from a bird's-eye view (BEV) of the learned representation and extract appearance embeddings for the next tracking stage.
In Section ~\ref{section:centerradarnet}, we detail each component, the training strategy, and the online tracking algorithm leveraging the learned appearance features.

CenterRadarNet is trained and evaluated on the K-Radar Dataset \cite{paek2023kradar}, which is the first autonomous vehicle dataset that provides 360-degree stereo camera images, LiDAR point clouds, and front-view 4D radar tensors for 3D object perception tasks.
With intensive experiments, CenterRadarNet surpasses previous methods on the K-Radar 3D object detection benchmark. 
Furthermore, we present the first tracking results on the K-Radar dataset V2.

\section{Related Work}
Radar has been applied in automotive for several decades, and many studies used radar to analyze the types and positions of objects that could appear on the road. 
Some works aim to detect Vulnerable Road Users (VRUs) \cite{VRUs-1}, while others focus on objects with potential risks to avoid collisions \cite{VehDet_CFAR1,CNN_RUDet_3DRadarCube,pre-crash_sensing}.
These applications require accurate 3D position and classification of objects to prevent accidents.

\begin{figure}[t]
    \centering
    \includegraphics[width=\linewidth]{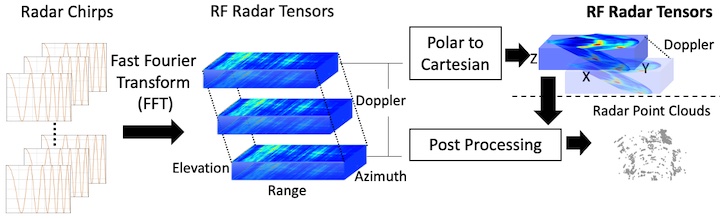}
    \caption{Radar Signal Processing and Representation. We use RF radar tensors after the polar to Cartesian coordinate transform as our model input.} 
    \label{fig:radar_process}
\end{figure}

Compared to cameras, radar has less semantic information \cite{Squeeze-and-Excitation} but provides precise positional information and is less susceptible to environmental factors such as weather conditions.
Thus, radar can be a good supplement to the camera as a more robust range-sensing modality.
Fig. \ref{fig:radar_process} shows two common radar data representations used in autonomous vehicles: RF radar tensors and radar point clouds. 
First, Fast Fourier Transform (FFT) is applied to radar's time series response and produces RF radar tensors.
Secondly, radar tensors undergo adaptive peak thresholding and clustering to produce radar point clouds. 
These thresholding and clustering mechanisms can lead to sparsity in the radar point clouds, resulting in information loss.

For radar perception tasks, some work proposed signal processing methods to extract features.
\cite{Yang2022CFAR} utilized cell averaging constant false alarm rate (CA-CFAR), which set an appropriate threshold, reducing false alarms and, as a result, improving the accuracy of radar systems.
\cite{Lin2016Springs} built a signal processing architecture and an algorithm to extract radar semantic features.
\cite{Malzer2021mdpi,Prophet2018irs} employed Density-Based Spatial Clustering of Applications (DBSCAN). 
\cite{Zhang2018radar} used Support Vector Machine (SVM) and \cite{scheiner:2019} used clustering approach for the object classification task.
These methods rely on human-engineered radar features and classifiers for the purpose of object classification.

Many studies have investigated the use of radar data for object detection.
Some approaches require post-processing radar data before feeding them into neural networks.
\cite{Patel2019CNN} combined CA-CFAR with a classifier based on a convolutional neural network (CNN).
\cite{Angelov2018IET} applied OS-CFAR and made a CNN-based LSTM architecture as a classifier.
However, recent research indicates that neural networks, when utilizing radar tensors that retain raw data, can offer more information without needing human-engineered post-processing.

\cite{CNN_RUDet_3DRadarCube} used a region-based CNN to detect and classify objects from RF radar data. 
\cite{wang2021rodnet2} proposed RODNet, which is supervised by point-level object labels provided by an automatic camera labeling pipeline.
\cite{Meyer2021CVF} constructed an isotropic graph convolutional neural network (GCN) to obtain the features and built a region proposal network (RPN) to get more information from different scales of RF radar tensors.
\cite{9022248} used a CNN that takes in range-azimuth-doppler radar tensors to infer objects' locations, sizes, and orientation in BEV.
\cite{9150751} introduces uncertainty estimate for object detection in BEV.
However, these methods only detect objects in BEV without the ability for object re-identification (re-ID).

Accurate object localization and estimation of size and rotation are pivotal for vehicular safety. CenterPoint \cite{CenterPoint} addresses this need with an efficient, anchor-free 3D object detection pipeline.

Multi-scale feature aggregation in the backbone allows the neural network to extract useful semantics and representations for perceiving objects at different scales. 
HRNet \cite{HRNet} excels in this context by offering enhanced feature learning through multi-scale representation aggregation. 
This makes it a versatile choice for various vision perception tasks.

%Tracking related work
With the advancement of object detection, tracking-by-detection has emerged as a predominant approach for object tracking. However, with a high dependency on detection results, tracking-by-detection is susceptible to failures when facing long missing tracks\cite{wang2021track}.

To address this, Joint Detection and Embedding (JDE) boosts the association by extracting embedding features of predicted bounding boxes simultaneously \cite{wang2020realtime,zhou2020tracking}. 
This approach allows them to leverage appearance information effectively, which in turn aids in addressing the challenge of object re-ID.
These trackers detect bounding boxes and associate them with the tracklets built by motion models  \cite{meinhardt2022trackformer,sun2021transtrack, zhou2020tracking, pang2021simpletrack}.

% Summarize
In summary, our work enhances 3D object detection by taking advantage of the rich feature set of RF radar tensors, which offer more informative characteristics than radar point clouds. To enhance the 3D object tracking, we employ JDE training across neighboring frames to capture discriminative appearance features for the re-ID task. 
These learned features are subsequently utilized in our online tracking algorithm.

\section{CenterRadarNet}
\label{section:centerradarnet}
CenterRadarNet is an identity-aware 3D object detection and tracking framework using 4D radar tensor as the sensor input.
It infers class-wise object existence's probability per BEV's grid, regresses 3D bounding parameters, and gathers every detected bounding box's appearance embedding in a single neural network's forward pass. 
The framework comprises a High-Resolution 3D backbone (HR3D), visibility-and-identity aware detection heads, and an online tracking algorithm leveraging the learned discriminative appearance embedding for bipartite matching. 

\subsection{High-Resolution 3D Backbone}
We build HR3D with fully 3D convolutional layers and treat the Doppler axis, \(D\), as the input channels.
As a result, each 3D convolutional kernel learns the volumetric spatial feature along \(Z, Y, X\) axes corresponding to the vertical, horizontal, and depth-wise directions in the Cartesian coordinate.
HR3D structure preserves spatial resolution and simultaneously learns useful semantics from radar's volumetric data by maintaining the highest-resolution representation and exchanging information with different-resolution features upon the switch of stages.
Fig. ~\ref{fig:cenrad} shows an example of HR3D structure with three stages. 
In the beginning, a series of convolutional layers transform the 4D radar tensors, \(T \in \mathbb{R}^{D \times Z \times Y \times X}\), into a feature, \(f_{in} \in \mathbb{R}^{C_{1} \times Z_{1} \times Y_{1} \times X_{1}}\).
From one stage to the next, the network grows another branch to expand the reception field by strided convolution. 
Each strided convolution transforms the input feature, \(f_{i} \in \mathbb{R}^{C_{i} \times Z_{i} \times Y_{i} \times X_{i}}\) , at branch \(i\)  to \(f_{i+1} \in \mathbb{R}^{ 2C_{i} \times \frac{Z_{i}}{2} \times \frac{Y_{i}}{2} \times \frac{X_{i}}{2} }\) at a new branch \(i+1\).
A stage comprises \(P\) consecutive modules that perform \(M\) times parallel convolution and exchange information between representations from different branches.
Following HRNet~\cite{HRNet}, we use either strided convolution or convolution followed by upsampling to match the feature dimensions from different branches during the information exchange.
We take the feature with the highest resolution and splat this feature into the BEV feature, \(f_{BEV} \in \mathbb{R}^{Y_{1} \times X_{1} \times C_{BEV}}\), by the reshape operation,  \(f \in \mathbb{R}^{C \times Z_{1} \times Y_{1} \times X_{1}} \Rightarrow f' \in \mathbb{R}^{C*Z_{1} \times Y_{1} \times X_{1}}\), and Conv2D layers.

\subsection{Visibility and Identity Aware Detection Head}
% Centerpoint detection head intro
Keypoint-based object detector \cite{centernet, cornernet} aim to identify the presence of objects per pixel from a shared dense feature map and decode the bounding box size by separate Conv2D heads at the same detected pixel coordinate. 
CenterPoint extends the keypoint-based learning to 3D detection by adding object rotation and height supervision.
HRNet backbone family\cite{higherhr, HRNet} has been proven to excel in keypoint detection and semantics learning.
Thus, we formalize our 3D object detection pipeline as the dense keypoint feature learning.
Following CenterPoint, we feed the BEV feature map into separate Conv.2D classification and regression heads.
Our preliminary study shows that the feature pyramid network before the heads presented in the CenterPoint is unnecessary due to HR3D's information exchange between low- and high-resolution representations.

We find that some labeled objects are hard to differentiate from the background noise by comparing the BEV radar tensor intensity change between regions enclosing bounding boxes and their surroundings.
This can be attributed to two main reasons. 
First, the much lower mounting position of the radar sensor than the LiDAR on a car makes the radar sensor occluded by high road concrete dividers.
Second, the polar-to-cartesian coordinate transform's interpolation causes increasing information loss for further objects.
To quantify each object's radar visibility, we analyze the radar tensors with a statistical CA-CFAR algorithm. The expression is simplified to 1-D kernel. We implement the algorithm with 3D kernel run on CUDA GPU and set \(N=15, G=5\) with voxel size of 0.4 meter.

\begin{algorithm}
\caption{Statistical CA-CFAR Algorithm}
\begin{algorithmic}[1]
\REQUIRE Array of power levels \( x \), Index of Cell Under Test \( c \), Number of training cells \( N \), Number of guard cells \( G \), Multiplier constants \( \alpha_1 \) and \( \alpha_2 \)
\ENSURE Threshold for detection at \( c \)
\STATE Compute the Average Noise Estimate:
\[
\text{AvgNoise} = \frac{1}{N} \sum_{i=c-G-N}^{c+G+N} x_i
\]
\STATE Compute the Standard Deviation of Noise:
\[
\text{StdNoise} = \sqrt{\frac{1}{N} \sum_{i=c-G-N}^{c+G+N} (x_i - \text{AvgNoise})^2}
\]
\STATE Compute the Modified CFAR Threshold:
\[
\text{Threshold} = \alpha_1 \times \text{AvgNoise} + \alpha_2 \times \text{StdNoise}
\]
\STATE \textbf{if} \( x_c > \text{Threshold} \) \textbf{then}
\STATE \quad Declare target present at \( c \)
\STATE \textbf{end if}
\end{algorithmic}
\end{algorithm}

\begin{figure}[!ht]
    \centering
    \includegraphics[width=1\linewidth]{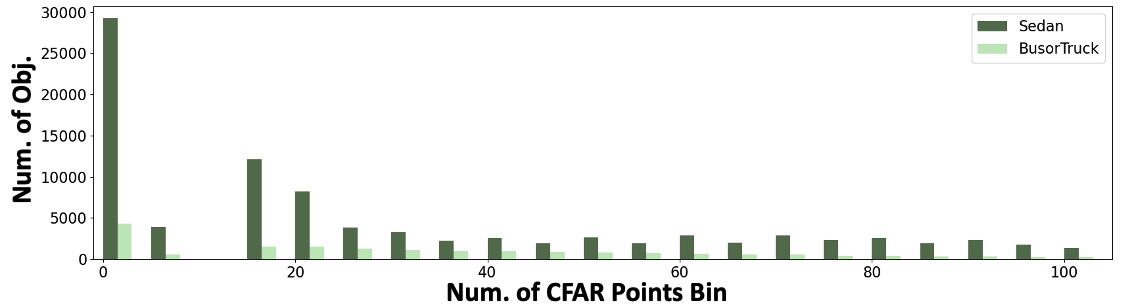}
    \caption{
        Histogram of numbers of CFAR points in objects from the K-Radar dataset. The bin is left-exclusive and right-inclusive with the size of 5. 
    }
    \label{fig:heads}
\end{figure}

The number of CFAR points in an object indicates the likelihood of an object feature being extracted from the background noise by the learned convolutional kernels.
We modify the ground truth heatmap, \(H\), preparation process from \cite{cornernet} to incorporate this statistical property into our object detection training strategy by soft labeling with Eq. \ref{eq:softlabel}, where \(\sigma_x=\frac{\alpha w_l}{6}, \sigma_y=\frac{\alpha h_l}{6}\). \(O\) and \(\alpha\) represent object and its size adaptive standard deviation respectively\cite{cornernet}. \(w()\) is the weighting function for soft labeling. 

\begin{equation}\label{eq:softlabel}
H(u, v, O) =  w(O) \exp \left( -\frac{(u-O_{cx})^2}{2\sigma_x^2} - \frac{(v-O_{cy})^2}{2\sigma_y^2} \right) 
\end{equation}

% appearance embedding head
CenterRadarNet extends the CenterPoint heads by adding a separate branch of Conv2D layers to decode the point feature of an object into the appearance embedding, \(emb_{App.}\in \mathbb{R}^{E})\).
We formalize our training as multi-task learning to jointly train the object detection and the appearance heads.
The next subsection introduces our joint classification, regression, and appearance heads training strategy.

\subsection{Cross-Frame Joint Detection and Appearance Training}
\begin{figure}[!ht]
    \centering
    \includegraphics[width=1\linewidth]{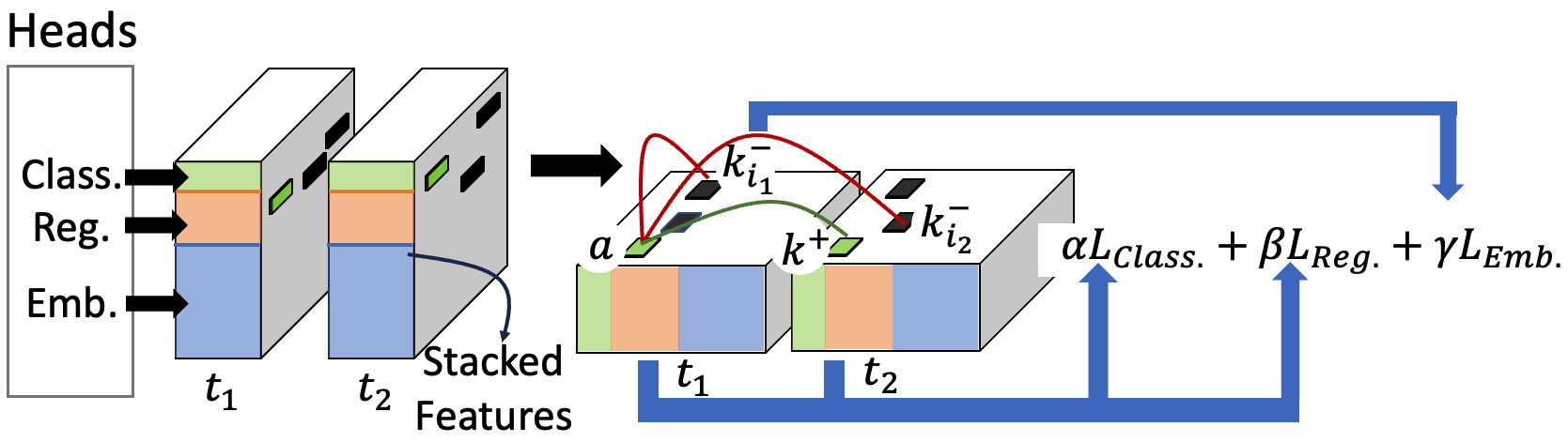}
    \caption{
        Joint detection and appearance embedding training procedure.
    }
    \label{fig:jde_trainng}
\end{figure}

% Training procedure
Fig. \ref{fig:jde_trainng} shows our joint detection and appearance feature training pipeline.
First, we sampled two frames at timestamp, \(t1\) and \(t2\), where \(|t1 - t2| < T\) from the training set to form each mini batch. Then, we feed their radar tensors into CenterRadarNet to get classification, regression, and appearance features. 
Second, we use classification and regression losses, \(\mathcal{L}_{Class.}\) and \(\mathcal{L}_{Reg.}\) from \cite{CenterPoint} for the 3D object detection supervision.
Finally, we adopt metric learning to supervise the appearance embedding head.
Fig. \ref{fig:jde_trainng} illustrates an example of anchor, positive and negatives. Two green boxes are objects at \(t1\) and \(t2\) with the same ground truth tracking ID while the black boxes belong to different tracking IDs.
We treat each object as one anchor, \(a\), the other object with the same ID as the positive, \(k^{+}\), and the remaining \(N\) objects (black boxes) as negatives, \(\{k^{-}_{i}\}_{i=1}^{N}\).
The appearance-embedding loss is chosen to be a triplet hard loss in Eq. \ref{eq:emb_loss}, where the distance function, \(\mathcal{D}\), compares the cosine similarity between two appearance embedding, \(\mathcal{F}(a)\) and \(\mathcal{F}(k)\). 
For each triplet, we mine the hard negative, \(k^{-}_{hard}\), by selecting the negative with the minimum distance to the anchor in Eq. \ref{eq:hard_mining}. \\
The multi-task learning results in the final loss in Eq. \ref{eq:final_loss}, where \(\alpha\), \(\beta\) and \(\gamma\) are constant coefficients.

\begin{gather} \label{eq:emb_loss}
    \mathcal{L}_{\text{Emb.}}(a, k^{+}, k^{-}_{\text{hard}}) = \\
    \max \left( 0, \mathcal{D}(\mathcal{F}(a), \mathcal{F}(k^{+})) - \mathcal{D}(\mathcal{F}(a), \mathcal{F}(k^{-}_{\text{hard}})) + m \right) \nonumber
\end{gather}

\begin{equation}\label{eq:hard_mining}
k^{-}_{hard} = \underset{\{k^{-}_{i}\}_{i=1}^{N}}{\arg\min} \, \mathcal{D}(a, k^-)
\end{equation}

\begin{equation}\label{eq:final_loss}
    \mathcal{L} = \alpha \mathcal{L}_{Class.} + \beta \mathcal{L}_{Reg.} + \gamma \mathcal{L}_{Emb.}
\end{equation}

\subsection{Online Tracking with ReID}
\begin{figure}[!ht]
    \centering
    \includegraphics[width=1\linewidth]{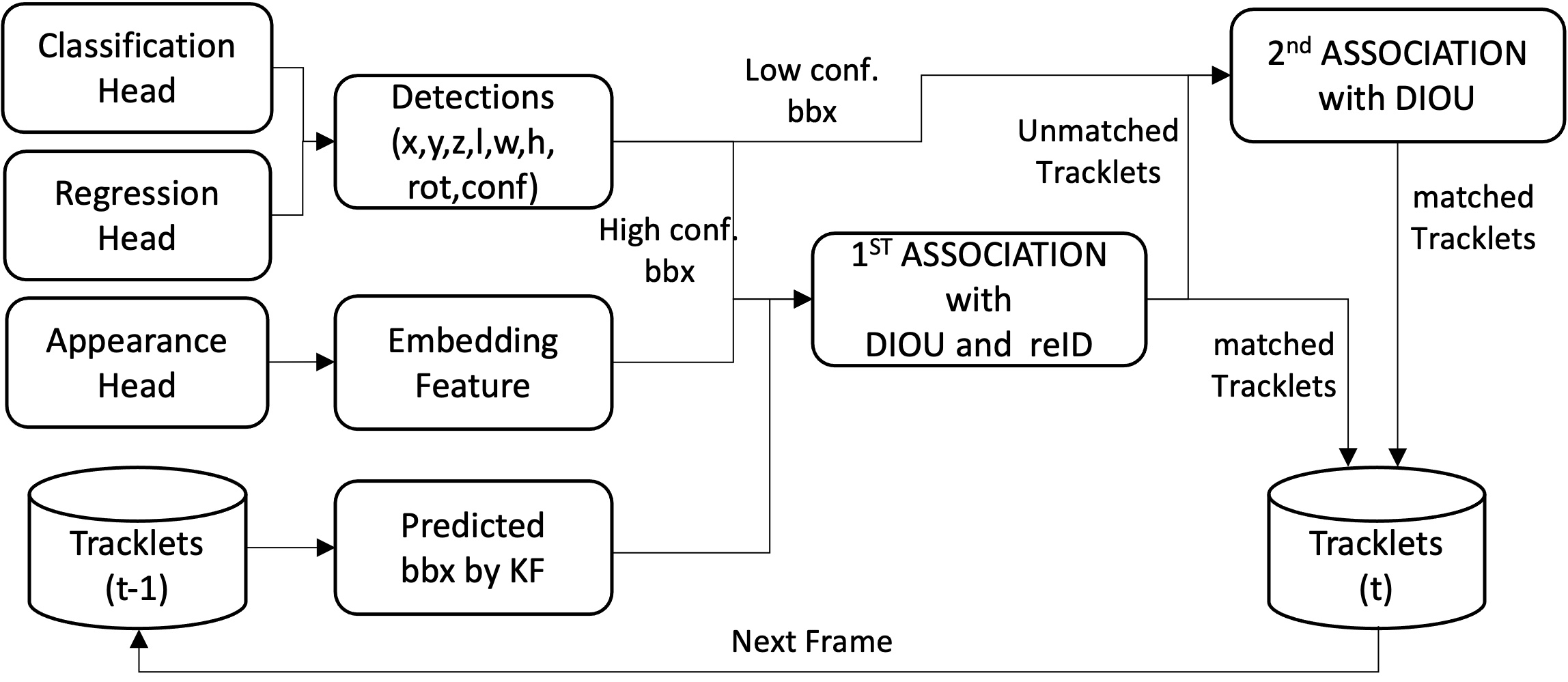}
    \caption{
        The pipeline of our online tracker. The state of detection is from the classification, regression heads and the embedding features are from the appearance head. The 'conf' in detection describes the confidence score of the bounding box, which is used to be the threshold of  'KF' means the Kalman filter and 'bbx' refers to the bounding boxes identified within a single frame.
    }
    \label{fig:OnlineTracker}
\end{figure}

To leverage the advancements in BEV representation, we incorporated bounding box information with appearance features into our online tracker.
We present our modifications and improvements to the tracking-by-detection methods by integrating these into  BoT-SorT\cite{aharon2022botsort}. The pipeline is presented in Fig. \ref{fig:OnlineTracker}. \\

% KF setting
To describe the object's motion, it is conventional to use the Kalman filter with a constant-velocity model. The goal is to estimate the state \(x \in \mathbb{R}^{n}\) with the given measurement \(z \in \mathbb{R}^{m}\). The Kalman filter is regulated by the Eq. \ref{eq:KFstate} and \ref{eq:KFmeasure}, where \(F_t\) is the transition matrix from \(t-1\) to \(t\) and \(H_t\) is the observation matrix. The \(n_t\) and \(v_t\) are the process and measurement noise.
\begin{equation}\label{eq:KFstate}
x_t= F_tx_{t-1} + n_{t-1} 
\end{equation}
\begin{equation}\label{eq:KFmeasure}
z_t = H_t x_t + v_t 
\end{equation}

% KF state setting
In the task of object tracking, we set the state vector by a sixteen-variable tuple, 
\begin{equation}\label{eq:KFstate}
x = [x,y,z,l,w,h,sin\theta,cos\theta,\dot{x},\dot{y},\dot{z},\dot{l},\dot{w},\dot{h},\dot{sin\theta},\dot{cos\theta}] 
\end{equation}
, where \((x_c,y_c,z_c)\) is the 3D coordinate of the bounding box center, \((l,w,h)\) is the scale of bounding box. \(\theta\) is the orientation along with axis z, so \((sin\theta,cos\theta)\) are used to describe the heading rotation.

%Describe the procedure of two association
Following ByteTrack\cite{zhang2022bytetrack} and BoT-SorT\cite{aharon2022botsort}, we leverage the presence of low score detection boxes by first pairing high score ones and subsequently associating them to the less confidence detection boxes. Since both bounding boxes and appearance features are generated simultaneously by the model, we only use appearance features when the confidence score of the bounding box exceeds a specific threshold.

When vehicles travel in the opposite lane or make turns at intersections, relying solely on the Intersection over Union (IoU) as the cost metric faces challenges in achieving accurate associations, given the vehicles' unpredictable and high speeds.
To address this, we replace \(IoU\) with \(DIoU\) in Eq.\ref{eq:DIOU} as the assignment cost\cite{zheng2019distanceiou}. \(DIoU\) minimizes the normalized distance between central points of two bounding boxes, \(b_1\) and \(b_2\). \(\rho(\cdot)\) is the Euclidean distance, and \(c\) is the diagonal length of the smallest enclosing box covering the two boxes.

\begin{equation}\label{eq:DIOU}
DIoU= 1 - IoU + \frac{\rho^2(b_1,b_2)}{c^2}
\end{equation}

\section{Experiments}
\label{section:Experiments}
\subsection{Dataset and Evaluation Metrics}
We evaluate CenterRadarNet on the K-Radar Dataset \cite{paek2023kradar} , which contains various driving scenarios with diverse illumination, weather conditions, and road structures.
It provides synchronized 64-channel and 128-channel LiDAR point clouds, 4D radar tensors, and 360-degree stereo images with objects' bounding box annotation relying on point clouds.
Among the raw continuous sequences, they split the whole dataset into around 17.5k training and 17.5k testing frames.
We adopt the K-Radar object detection evaluation metric and benchmark the 3D and BEV detection accuracy, AP\(^{3D}\), AP\(^{BEV}\), at the IoU threshold of 0.3.
We report all the results based on a multi-class detection setting and choose the two major classes, 'Sedan' and 'Bus or Truck,' as our detection targets.
We filter out objects with zero CFAR points in our training set to stabilize the training.
Since the LiDAR and the front camera are totally occluded in sequences 51, 52, 57, and 58, we do not use these sequences in our training and testing set.

In the evaluation for tracking, we use MOTA \cite{Bernardin2008} and IDF1\cite{ristani2016performance} to show the performance of detection and association respectively.
To optimize the dataset for both detection and tracking tasks, we have enhanced the K-Radar dataset. This involves re-annotating each bounding box and supplementing missing tracking IDs for all sequences. We make our annotations publicly available\footnote{Our annotations can be accessed at:\\ \url{https://shorturl.at/hmFG8}}.

\subsection{Implementation}
We used the 4D radar tensors provided by the dataset and converted them from the polar coordinate to the Cartesian coordinate.
Each voxel grid in a radar tensor occupies the metric size of (0.4m, 0.4m, 0.4m) in length, width, and height according to the resolution of the radar sensor.
In CenterRadarNet, we configured HR3D with \((N,P,M)=(3, 1, 2)\), and the maximum number of objects is set to 30 in the detection head due to fewer objects appearing in a time frame than other public datasets.
To compare with baseline methods, we implemented CenterPoint with backbones from PointPillar\cite{lang2019pointpillars} and VoxelNet\cite{zhou2017voxelnet}, taking in radar point clouds from our CFAR algorithm.
For the JDE, we set \(T=5\) and the dimension of appearance embedding to be 32.
We set the detection range, \((x_{min}, x_{max}, y_{min}, y_{max}, z_{min}, z_{max})=(0, 72, -15, 15, -2, 7.6)\) meters, which widens the detection range compared the original setup in \cite{paek2023kradar}.
We trained all the experiments using a V100 GPU with a batch size of 16 for 30 epochs.
All the evaluations were done on a 3090 GPU.

\subsection{Results}
Table \ref{tab:exp_overall} shows that CenterRadarNet surpasses baseline methods using 4D radar in terms of multi-class AP\(^{3D}\) and AP\(^{BEV}\) and achieves the stat-of-the-art performance on the K-Radar 3D object detection benchmark. 
Evaluation, as detailed in \cite{paek2023kradar}, relies on the predictions of a single-class model.
The results exhibit the superiority of CenterRadarNet to extract high-resolution representations over the baseline methods, which first apply signal processing on radar tensors and then feed into their neural networks. 
The heuristic selection of radar features in the signal processing stage may not achieve the optimal feature extraction. 
Instead, our model bypasses the first signal processing stage and learns to extract useful features from the raw radar tensors.

\begin{table}[h!]
    \caption{Performance comparison of 3D object detection in Kradar}
    \label{tab:exp_overall}
    \centering
    \small
    \setlength\tabcolsep{2.5pt}{
    \begin{tabular}{c|cc|cc}
        \toprule
        \multirow{2}{*}{Method} & \multicolumn{2}{c|}{Sedan} & \multicolumn{2}{c}{Bus or Truck} \\
          & AP$^{3D}$ & AP$^{BEV}$ & AP$^{3D}$ & AP$^{BEV}$ \\
        \midrule
        \midrule
        RTNH \cite{paek2023kradar}& 47.44 & 58.39 & NA & NA \\
        CenterPoint(PointPiller)\cite{lang2019pointpillars} & 31.36 & 34.48 & 17.11 & 19.44 \\
        CenterPoint(VoxelNet)\cite{zhou2017voxelnet} & 50.83 & 54.93 & 27.44 & 29.48 \\
        \midrule
        \midrule

        \textbf{CenterRadarNet} & \textbf{55.36} & \textbf{64.03} & \textbf{28.49} & \textbf{30.23} \\
        \bottomrule
    \end{tabular}}
\end{table}

% CenterRadarNet module design
To systematically assess the impact of each design element, we performed an ablation study comparing our CenterRadarNet framework with various alternative configurations. When Conv3D layers are replaced by Conv2D in the HR3D, there is a noticeable decline in detection performance, which can be attributed to the loss of spatial information in the elevation axis. 
The information exchange mechanism, aggregating features across different levels, notably enhances the detection of large objects.
Lastly, we demonstrate that the inclusion of JDE training in a multi-task learning setup does not compromise the detection performance. On the contrary, it enables the model to generate more discriminative appearance embedding, which are crucial for effective online tracking in the subsequent tracking stage.

\begin{table}[h!]
    \caption{Ablation analysis on CenterRadarNet configurations}
    \label{tab:ablation}
    \centering
    \small
    \setlength\tabcolsep{1pt}{
    \begin{tabular}{c|c|c|cc|cc}
        \toprule
        \multirow{2}{*}{ \makecell[c]{Info. \\Exchange}} & \multirow{2}{*}{\makecell[c]{3D \\Backbone}}  & \makecell[c]{\\JDE} & \multicolumn{2}{c|}{Sedan} & \multicolumn{2}{c}{Bus or Truck} \\
          &&& AP$^{3D}$ & AP$^{BEV}$ & AP$^{3D}$ & AP$^{BEV}$ \\
        \midrule
        \midrule
        V&&&  46.71 & 55.16 & 26.28 & 29.83\\
        &V&&  53.36 & 61.56 & 20.37 & 23.84\\
        % V&V&&&  54.95 & 57.78 & 29.17 & 31.02\\
        V&V&& 55.24 & \textbf{64.51} & 28.04 & 30.10\\
        \midrule
        V&V&V & \textbf{55.36} & 64.03 & \textbf{28.49} & \textbf{30.23}\\
        \bottomrule
    \end{tabular}}
\end{table}

% JDE loss
We conduct an ablation study on the choice of loss function employed in learning the appearance embedding. 
The methods under comparison employ identical configurations for both the HR3D backbone and the associated heads.
Our findings indicate that when paired with cosine similarity as the distance function, triplet loss performs better than InfoNCE~\cite{infonce} and triplet loss using Euclidean distance.
Subsequently, we evaluate these JDE methods within the context of the object-tracking task.

\begin{table}[h!]
    \caption{JDE Loss }
    \label{tab:JDE_loss}
    \centering
    \small
    
    \begin{tabular}{c|cc|cc}
        \toprule
        \multirow{2}{*}{Loss} & \multicolumn{2}{c|}{Sedan} & \multicolumn{2}{c}{Bus or Truck} \\
          & AP$^{3D}$ & AP$^{BEV}$ & AP$^{3D}$ & AP$^{BEV}$ \\
        \midrule
        \midrule
        InfoNCE & 48.71 & 58.10 & 28.48 & \textbf{31.43} \\
        \makecell[c]{Triplet\\(L2)}& 55.07 & 57.68 & 27.83 & 30.51  \\
        \makecell[c]{Triplet\\(Cosine Similarity)}& \textbf{55.36} & \textbf{64.03} & \textbf{28.49} & 30.23 \\
        \bottomrule
    \end{tabular}
\end{table}

% Online tracker
We compared the tracking performance of different losses used in the JDE training and different cost functions for the association. 
In Table \ref{tab:trackerPerformance}, we found that using DIoU as the assignment cost outperforms using IoU in all scenarios.

\begin{figure}[!ht]
    \centering
    \includegraphics[width=1\linewidth]{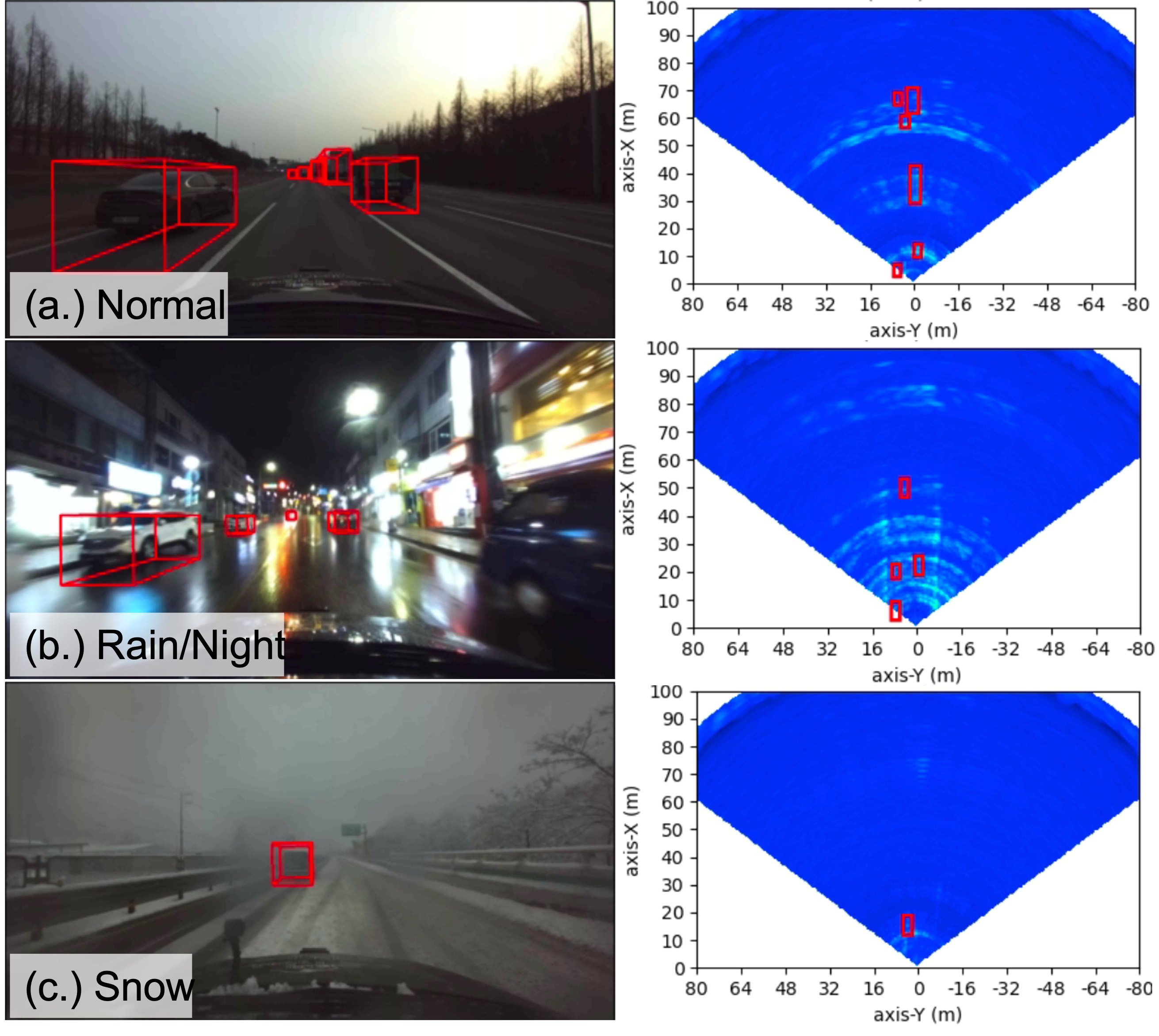}
    \caption{
    CenterRadarNet's prediction results on the K-Radar test set are visualized in both the front camera's field of view and the radar Bird's Eye View (BEV) with power intensity serving as the background. The successful detection of objects under various weather conditions attests to the robustness of using radar as the sensor input.
    }
    \label{fig:example_viz}
\end{figure}

Fig. ~\ref{fig:example_viz} illustrates examples of 3D bounding boxes predicted by CenterRadarNet. These are visualized in the front camera view and the radar BEV.
We provide a demo video to demonstrate CenterRadarNet's qualitative results of 3D object detection and tracking\footnote{Demo of CenterRadarNet's object detection and tracking using only FMCW radar: \url{https://youtu.be/Z5Kxn_uyHLM}}.
To substantiate the feasibility of employing CenterRadarNet in real-time applications, we report that our detection and tracking pipeline operates at 23 frames per second (FPS) on a 3090 GPU.

\subsection{Limitations}
\begin{figure}[!ht]
    \centering
    \includegraphics[width=1\linewidth]{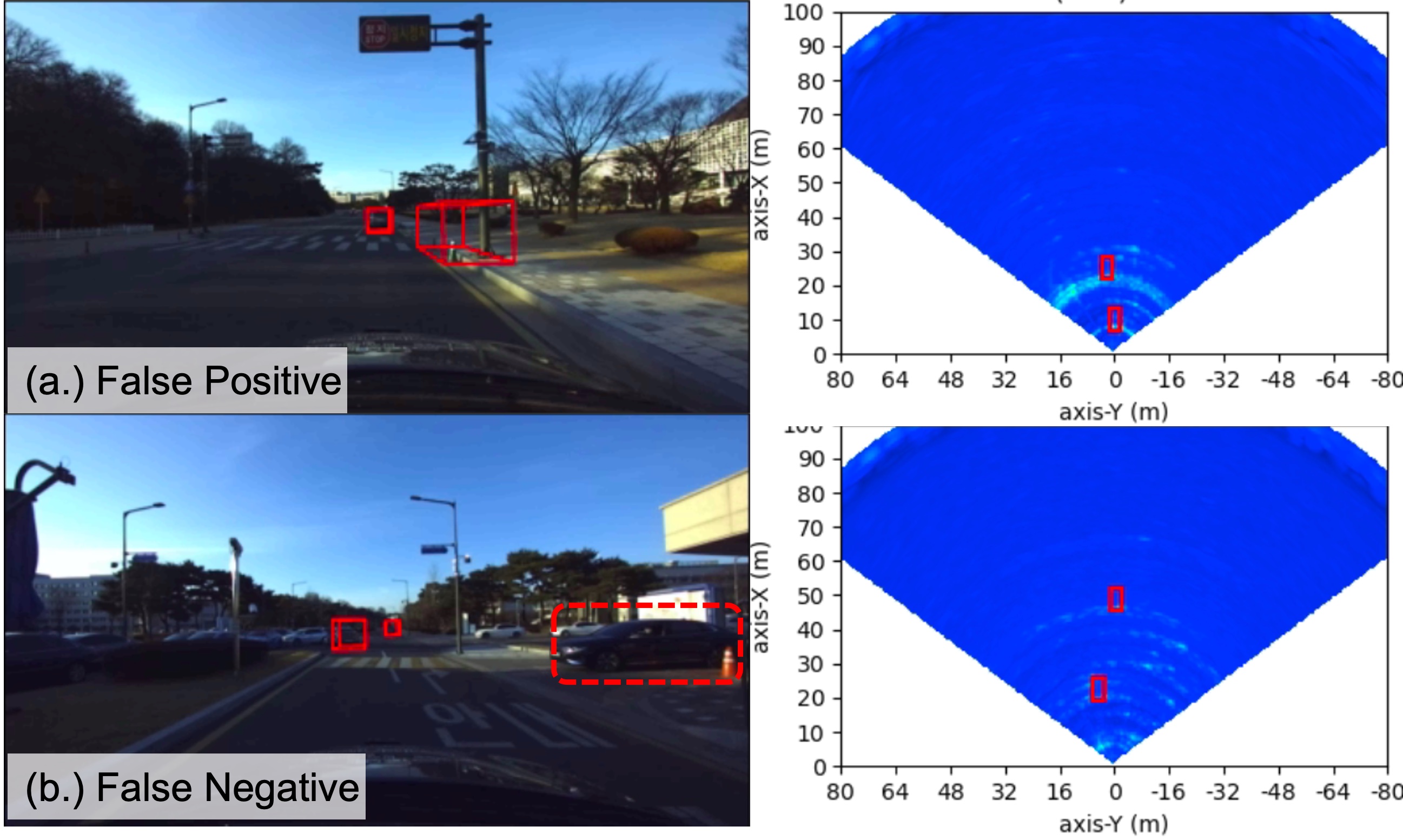}
    \caption{(a.) shows common false positive cases on the roadside objects. (b.) shows an example of false negative detection of the vehicle.
    }
    \label{fig:Limitations}
\end{figure}

In the context of some particular road scenario, our current finding still has several limitations. In scenes with many diverse and complex objects, especially containing metallic materials, it often leads to false positive detection by our model, as shown in Fig. \ref{fig:Limitations}(a). By expanding the scope of object recognition to encompass additional common categories on roadways may enhance the results. In addition, we found that when the vehicle's travel direction is not parallel to the ego car, the recognition capability of our current model is also insufficient, as shown in Fig. \ref{fig:Limitations}(b). 
We attribute this issue to the absence of such vehicles in K-Radar dataset, and we anticipate potential improvement in our results with the availability of more comprehensive data in the future. 

\begin{table}[!h]
    \caption{Online Tracker Performance}
    \label{tab:trackerPerformance}
    \centering
    \small
    \setlength\tabcolsep{3pt}{
    \begin{tabular}{c|c|c|c|c|c|c}
        \toprule
        
        \makecell[c]{Loss\\ Method}&\makecell[c]{Cost \\Func. }& \makecell[c]{MOTA\\(\%)} & \makecell[c]{IDF1\\ (\%)}&FP&FN&IDs\\
        \midrule
        \midrule
        \multirow{2}{*}{InfoNCE}&IoU & 43.3 &59.2  & 1,509 & 12,902& 412 \\
        & DIoU & 44.1& 61.3 & 1,507 & 12,807& 297 \\
        \midrule
        \multirowcell{2}{Triplet\\(L2)}&IoU & 47.4& 63.6 & 2,222 & 11,101& 424 \\
        & DIoU &47.5 & 64.3 & 2,302 & 11,042& 372 \\
        \midrule
        \multirowcell{2}{Triplet\\(Cosine Similarity)}&IoU & 47.7 &63.0 & 1,960  & 11,239& 489 \\
        & DIoU & \textbf{48.7} & \textbf{65.2} & 1,943 & 11,150& 315 \\

        \bottomrule
    \end{tabular}}
\end{table}

\section{Conclusion}
We introduce CenterRadarNet, an online detection-and-tracking architecture using 4D RF radar tensors to perform 3D object detection and tracking tasks. 
CenterRadarNet is tailored to joint learning for object detection and appearance embedding, which advances re-ID capabilities. 
Our method not only achieves the state-of-the-art result in the K-Radar 3D object detection benchmark but also pioneers the establishment of the first object-tracking result on the K-Radar dataset V2. 
In testing under diverse driving scenarios, CenterRadarNet demonstrates consistent performance, underscoring its applicability across varied driving conditions.

{
\small
\bibliographystyle{plain}
\bibliography{ref}
}

\end{document}